\documentclass[letterpaper,10pt]{article} 

\usepackage{osameet3} 

\usepackage{amsmath,amssymb}
\usepackage[colorlinks=true,bookmarks=false,citecolor=blue,urlcolor=blue]{hyperref} 

\begin{document}




\title{Comparison of Models for Training Optical Matrix Multipliers in Neuromorphic PICs}

\author{A. Cem\textsuperscript{1*}, S. Yan\textsuperscript{1,2}, U.C. de Moura\textsuperscript{1}, Y. Ding\textsuperscript{1}, D. Zibar\textsuperscript{1}, F. Da Ros\textsuperscript{1}}
\address{\textsuperscript{1}DTU Fotonik, Technical University of Denmark, DK-2800, Kgs. Lyngby, Denmark \\
\textsuperscript{2}School of Optical \& Electrical Information, Huazhong Univ. of Science and Technology, 430074, Wuhan, China}
\email{*alice@dtu.dk}
\copyrightyear{2021}
\begin{abstract}
We experimentally compare simple physics-based vs. data-driven neural-network-based models for offline training of programmable photonic chips using Mach-Zehnder interferometer meshes. The neural-network model outperforms physics-based models for a chip with thermal crosstalk, yielding increased testing accuracy. 
\end{abstract}
\section{Introduction}
The increasing demand for computational power, especially for machine learning applications, has resulted in a renewed interest in the development of new hardware architectures. Particularly, photonics has emerged as a promising medium for neuromorphic computing \cite{shastri2021}. Motivated by the impressive results obtained using artificial neural networks (ANNs), optical neural networks (ONNs) have been suggested as a fast and low-power alternative. Each layer of an ANN consists of two fundamental operations: a linear matrix-vector multiplication followed by a nonlinear activation function. Photonics implements the former particularly efficiently with a variety of designs proposed, including using tunable microring resonators \cite{tait2017} and Mach-Zehnder interferometer (MZI) meshes \cite{shen2017}.

Focusing on the latter, the linear weights are typically adjusted by changing the voltages applied to the thermo-optic phase shifters present in each MZI. In order to program a MZI mesh to implement a weight matrix, the mapping between the input voltages and MZI transmission is required. This can be obtained using the simple physics-based model of a MZI \cite{shen2017,perez2017}. Due to fabrication errors and neglected effects, such as thermal crosstalk (XT), using the simple physics-based model results in poor performance~\cite{perez2017,bandyopadhyay2021,fang2019}. This has lead to the development of either error correction schemes \cite{bandyopadhyay2021} and error-resistant ONNs \cite{fang2019}, or calibration procedures characterizing the voltage-transmission relation of each MZI individually~\cite{perez2017}. However, the calibration of~\cite{perez2017} does not take into account the thermal XT between MZIs. This is an effect that increases as the number of MZIs per chip footprint grows, even though complex crosstalk-minimizing chip designs \cite{milanizadeh2020} are being proposed. On-line training procedures~\cite{zhang2021} address this challenge but require re-optimization for each new target, thus they do not remove the need for accurate models.

In this work, we compare the performances of simple physics-based models with and without thermal XT versus an ANN approach for modelling MZI meshes, i.e. ONN matrix multipliers. All models are trained with experimental data to predict the weights of the MZI mesh given the voltages applied to the individual MZIs for phase tuning. We evaluate the models using the photonic integrated circuit (PIC) described in \cite{ding2016}. The ANN model is significantly more accurate than simple physics-based models and results in improved testing performance when used to numerically validate the training for solving a toy classification task.
\vspace{-0.1cm}

\section{MZI Mesh Models}
\vspace{-0.3cm}
\begin{figure}[htbp]
  \centering
  \includegraphics[width=0.95\textwidth]{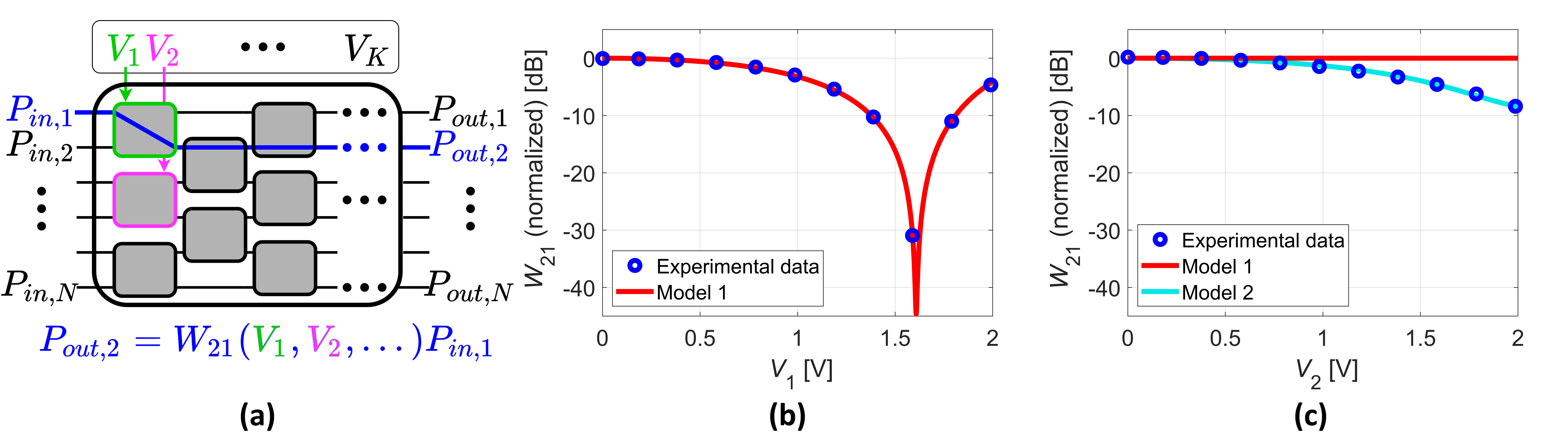}
\caption{(a) Block diagram of a photonic matrix multiplier. Experimental measurements for $W_{21}$ when the (b) $V_1$ and (c) $V_2$ are swept, plotted along the fitted curves for the physics-based models.}\vspace{-.6cm}
\label{fig1}
\end{figure}

An example of an architecture for a MZI mesh implementing the linear layer of an ONN is shown in Fig. \ref{fig1}(a). Each MZI has a thermo-optic phase shifter controlled by an external heater voltage on one arm, which can be tuned to modify the implemented weight matrix $\mathbf{W}$, relating the input and output power vectors $\mathbf{P_{in}}$ and $\mathbf{P_{out}}$ as follows: $\mathbf{P_{out}} = \mathbf{W}(\mathbf{V}) \cdot \mathbf{P_{in}}$. In order to program the PIC to implement a given matrix multiplication, a model describing the mapping between the heater voltages $\mathbf{V}$ and the achieved $\mathbf{W}$ is required.

We consider three models: two simple physics-based models without (Model 1) and with (Model 2) thermal XT and a data-driven ANN model (Model 3).
For Model 1, we have modified the model of a single MZI used in \cite{shen2017,perez2017} by adding a finite extinction ratio $ER$. The model for the weights $W_{ij}$ of the MZI mesh is given by Eq.~\ref{eqn1}. 
\begin{eqnarray}
W_{ij} = L_{ij} \prod_{k \in K_{ij}} \frac{1}{4} \left| \frac{\sqrt{ER} - 1}{\sqrt{ER} + 1} - \exp \left({\sqrt{-1} \; ( \phi^{(0)}_k + \phi^{(2)}_k V_k^2)} \right) \right|^2\,,
\label{eqn1}
\end{eqnarray}
where $L_{ij}$ is the total loss accumulated for the path through the PIC corresponding to $W_{ij}$, $K_{ij}$ is the set of all MZIs in path $i\rightarrow j$, $V_k$ is the voltage applied to the $k$\textsuperscript{th} MZI, and $\phi_k^{(0)}$ and $\phi_k^{(2)}$ are the fitted phase parameters describing the initial value of the phase offset and the quadratic (thermal) relation between $V_k$ and phase shift, respectively. Parameters $L_{ij}$, $\phi_k^{(0)}$ and $\phi_k^{(2)}$ are susceptible to fabrication errors and noise sources, so these values can be fitted during calibration, e.g. the voltage to one MZI is swept at a time and the output power is measured. This is equivalent to measuring $W_{ij}$ after input normalization. Fig. \ref{fig1}(b) shows an example sweep along with Model 1 trained by minimizing the root mean squared error (RMSE) between the experimentally measured and the predicted matrix weights in dB. RMSE is the cost function used for training all models. Even if the experimental data fits well to the analytical expression for this case, this does not necessarily mean that the chip is modelled accurately using Model 1. The thermal effect also applies undesired phase shifts to neighboring MZIs due to thermal XT, as illustrated in Fig. \ref{fig1}(c), resulting in errors for Model 1 when MZIs outside the path are controlled. Without characterizing the XT explicitly, Model 1 can be extended to capture this effect by including contributions from all voltages for the phase shifts, here called Model 2. $W_{ij}$ for a mesh with $M$ MZIs is given in Eq. \ref{eqn2}. Note that $\phi^{(2)}_{ij,k,m}$ denotes the impact of $V_m$ on MZI $k$ for the path corresponding to $W_{ij}$. Phase parameters were recalculated for each $W_{ij}$.
\begin{eqnarray}
W_{ij} = L_{ij} \prod_{k \in K_{ij}} \frac{1}{4} \left| \frac{\sqrt{ER} - 1}{\sqrt{ER} + 1} - \exp \left({\sqrt{-1} \; ( \phi^{(0)}_{ij,k} + \sum_{m=1}^M \phi^{(2)}_{ij,k,m} V_m^2)} \right) \right|^2\,.
\label{eqn2}
\end{eqnarray}
Alternatively, an ANN can be used to model the relation between $\mathbf{V}$ and $\mathbf{W}$ as shown in Fig. \ref{fig2}(a), here called Model 3. Getting inspiration from the physics-based models, we have included the heater voltages along with their squares in the input layer, both sets normalized between -1 and +1 separately. The output layer consists of the matrix weights in dB. To train an accurate model, an experimental dataset consisting of many different combinations of input voltages is required. ANN hyperparameters are optimized such that the RMSE between the measured and the predicted outputs is minimized, resulting in the architecture shown in Fig. \ref{fig2}(a). The hyperbolic tangent was used as the nonlinear activation function in the hidden nodes.

\vspace{-.2cm}
\section{Experimental Setup and Results}
\begin{figure}[htbp]
  \centering
  \vspace{-.2cm}\includegraphics[width=0.95\textwidth]{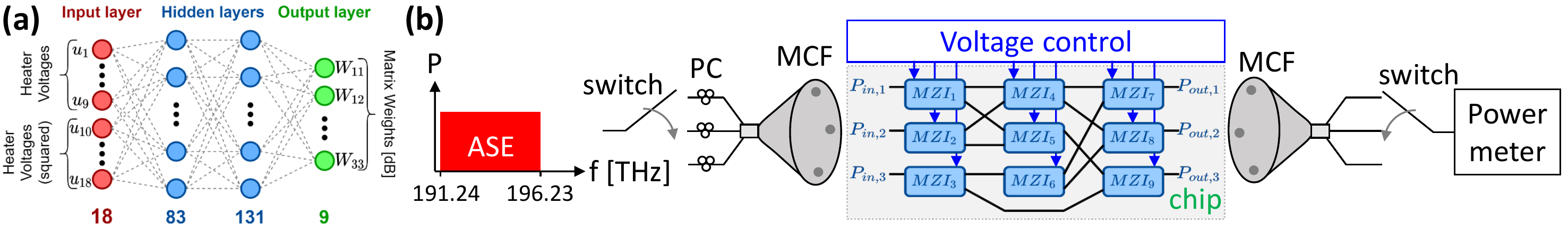}
\caption{(a) Architecture of the ANN model (Model 3) for the PIC. (b) Measurement setup for acquiring the training and testing datasets from the PIC. }
\label{fig2}
\end{figure}

\vspace{-.3cm}
The measurement setup is shown in Fig. \ref{fig2}(b).
A flattened power spectral distribution from an amplified spontaneous emission (ASE) source (5-THz bandwidth) is injected to the chip. The light is coupled from the three-core multi-core fiber (MCF) to the chip by an on-chip grating coupler array\cite{ding2016}. Polarization controllers (PC) control the input signal polarization, and $1\times3$ and $3\times1$ optical switches are used at the input and output of the chip to measure the $3\times3$ weight matrix using a power meter. The details of the photonic chip considered for the measurements are discussed in \cite{ding2016}. Here, we considered only a subset of the MZIs that resulted in a $3\times3$ weight matrix (chip in Fig. \ref{fig2}(b)), with their voltages independently controlled. 
To acquire a measurement dataset so that the three models can be compared, control voltages were individually varied from 0 to 2V (half-period of the MZIs), either systematically and one at a time (for fitting the phase parameters of Model 1) or measuring up to 5100 sets of ($\mathbf{V}$, $\mathbf{W}$) by randomly varying the applied voltages (for fitting the losses in Model 1 and fitting all parameters in Models 2 and 3). 4400 measurement samples were used for training, and the remaining 700 as testing data. For Model 3, 700 samples from the training set were reserved for cross-validation. Models 1 and 2 were trained on MATLAB with the interior-point algorithm, while Model 3 was trained on PyTorch with the L-BFGS optimizer.

\begin{figure}[htbp]
  \centering
  \includegraphics[width=0.95\textwidth]{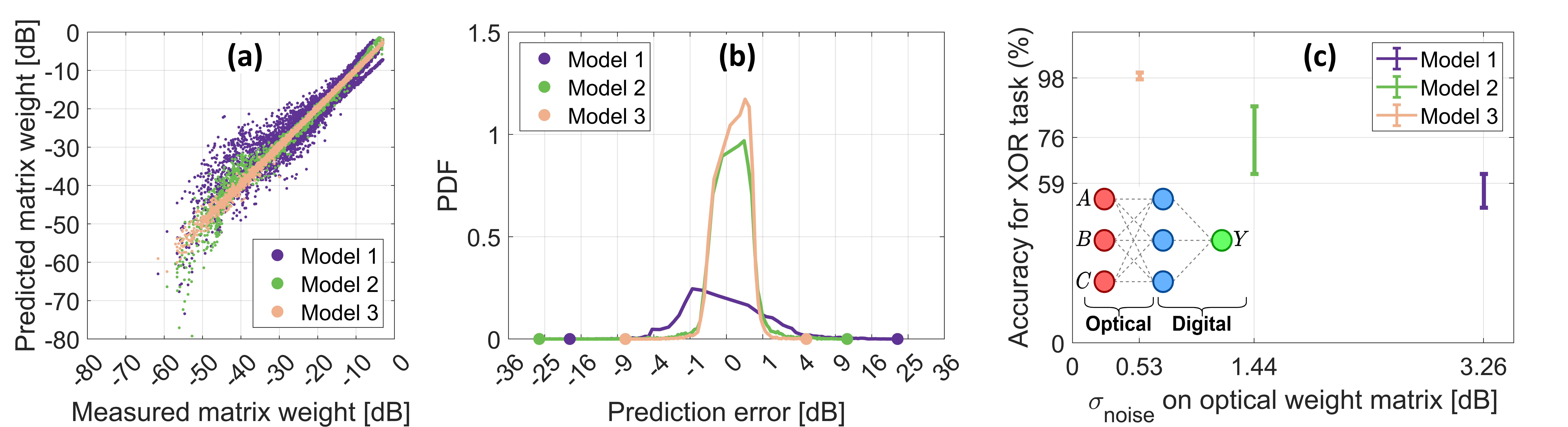}
\caption{(a) Scatter plot of the predicted and measured matrix weights for the three models. (b) Probability density functions obtained by normalizing the error histograms. (c) Classification accuracy on the 3-input XOR task with noise ($\sigma_{noise}$) added on $\mathbf{W}$  according to the training model's RMSE. Error bars show the 25\textsuperscript{th} and 75\textsuperscript{th} percentiles for 20 noise realizations. Inset: architecture of the ANN.}
\label{fig3}
\end{figure}

After training, the RMSEs between the measured and the predicted matrix weights in the testing set are 3.26 dB, 1.44 dB, and 0.53 dB for Models 1, 2, and 3, respectively. Fig. \ref{fig3}(a) shows the scatter plots of all weights for the testing data, revealing that Model 3 achieves superior performance both for smaller and larger weights. The probability distribution functions (PDFs) of the errors between predicted and measured weights over the testing data are plotted in Fig. \ref{fig3}(b), for all three models. While considering the thermal XT in the physics-based model (Model 2) results in a narrower error distribution, Models 1 and 2 are not as accurate as  Model 3, especially in terms of maximum prediction error, resulting in the long tails for their respective PDFs.

To quantify the advantage of more accurate models for off-line training in a practical case, an example single-hidden-layer ANN was trained digitally for the 3-input-bit XOR task described in \cite{williamson2020} (output is 1 only when exactly one input is 1). We assumed that the MZI mesh was used as the $3\times3$ matrix of weights between the input and the hidden layer (inset Fig.~\ref{fig3}(c)). After training, Gaussian noise was introduced to the weights of the $3\times3$ linear layer in order to quantify the accuracy penalty that comes from using an imperfect model for training, i.e. the applied voltages yield slightly different weights. The noise effectively emulates prediction errors between the training model and the chip. The hidden nodes with sigmoid activation functions and the $3\times1$ linear output layer were kept noise-free. The classification accuracies for the noisy neural networks are shown in Fig. \ref{fig3}(c). Using Model 3 (RMSE = 0.59 dB) for training yields higher testing accuracy ($>98$\% for 20 different error realizations) than by considering the physics-based models (76\% and 59\% accuracy for Models 2 and 1, respectively).
\vspace{-.2cm}
\vspace{-.6cm}
\section{Conclusion}
We describe and experimentally test an ANN-based approach for modelling photonic neuromorphic devices. Compared to the simple physics-based models, the ANN-based model reduces the prediction error by more than a factor of 2 for a fabricated chip, even when the thermal XT is accounted for by the physics-based model. While our approach was tested on a specific chip, it can also be applied to other more complex MZI-mesh architectures.

\small{\noindent\textbf{Acknowledgment} Villum Foundations, Villum YI, OPTIC-AI, grant n. 29344, and ERC CoG FRECOM, grant n. 771878.}

\end{document}